\documentclass{Interspeech2024}
\usepackage{enumerate}
\usepackage{inputenc}
\usepackage{multirow}
\usepackage{longtable}
\usepackage{geometry}
\usepackage{booktabs}
\usepackage{graphicx}
\usepackage{booktabs}
\usepackage{caption}
\usepackage{makecell}

\interspeechcameraready

\title{Decoding Emotion: Speech Perception Patterns in Individuals with Self-reported Depression}

\name[affiliation={1, 2}]{Guneesh}{Vats}
\name[affiliation={1}]{Priyanka}{Srivastava}
\name[affiliation={1, 2}]{Chiranjeevi}{Yarra}

\address{
  $^1$International Institute of Information Technology(IIIT), Hyderabad, India\\
  $^2$Speech Processing Lab, IIIT Hyderabad} 
\email{guneesh.vats@research.iiit.ac.in, priyanka.srivastava@iiit.ac.in, chiranjeevi.yarra@iiit.ac.in}

\keywords{depression, affective speech perception, emotional reactivity, positive and negative scale (PANAS), Beck Depression Model, ECI model of depression}

\begin{document}

\maketitle

\begin{abstract}
    
The current study examines the relationship between self-reported depression and the perception of affective speech within the Indian population. PANAS and PHQ-9 were used to assess current mood and depression, respectively. Participants' emotional reactivity was recorded on a valence and arousal scale against the affective speech audio presented in a sequence. No significant differences between the depression and no-depression groups were observed for any of the emotional stimuli, except the audio file depicting neutral emotion. Significantly higher PANAS scores by the depression than the no-depression group indicate the impact of pre-disposed mood on the current mood status. Contrary to previous findings, this study did not observe reduced positive emotional reactivity by the depression group. However, the results demonstrated consistency in emotional reactivity for speech stimuli depicting sadness and anger across all measures of emotion perception.

\end{abstract}

\section{Introduction}

Speech perception plays a fundamental role in human communication and facilitates social interaction by analyzing the expression of complex thoughts and emotions. How we perceive speech depends not only on linguistic and acoustic features (e.g., words, and fundamental frequency), its content, the affective nature (e.g., anger or happy) or the paralinguistic cues (e.g., pause) but is also affected by the listeners’ mood. While extensive research has investigated the role of mood in processing emotional stimuli using visual stimuli either static or dynamic [4], words [1] or music [6], only a few studies have examined its role in speech affective perception [7, 8, 9, 10]. Noteworthy, the representation of Asian population, more specifically, Indian population has been dismal among these studies \cite{uekermann2008perception, peron2011major, punkanen2011biased} resulting in poor understanding of role of mood in speech perception from the Indian perspectives. We aim to address this gap by investigating the impact of predisposed mood, like depressive state, on affective speech perception by analyzing emotional reaction to the affective speech stimuli among the Indian population.

Mood becomes a relevant component for emotional reaction when it is characterized by persistent sadness and/ or loss of interest and pleasure in regular everyday activities, indicative of a clinical condition called depression or unipolar depression. Depression has been associated with impaired emotional regulation alongside aberration in somatic, cognitive, and psychomotor responses \cite{Sheoran2022, First2013}. In such cases, measure of emotional reactivity to positive and negative affective stimuli serves as a hallmark for examining affective processing related to an individual's depressive state \cite{Sheoran2022, Bylsma2008, koch2018neural, schlipf2013judgment}.
Emotional reactivity is defined as a change in individuals’ affective perception or experience in correspondence with the emotionally evocative stimuli \cite{Sheoran2022, Bylsma2008}. Despite common assumptions about reduced positive emotional reactivity \cite{Bylsma2008, Beck2008}, and increased negative emotional reactivity during depressive episodes \cite{Beck2008},  cognitive models \cite{ Bylsma2008, Beck2008}, and empirical studies \cite{uekermann2008perception, schlipf2013judgment} present contradictory views.

Recent studies \cite{uekermann2008perception,  koch2018neural, schlipf2013judgment},  using prosody, showed attenuated positive emotional reactivity (PER) to speech depicting happy emotion by the individuals experiencing depressive episode, whereas speech depicting angry and neutral or sad emotion did not elicit a significantly different emotional reaction between depression and no-depression groups  \cite{uekermann2008perception, schlipf2013judgment}. Similar results were observed in a study using fMRI technique with no difference in amygdala activity was observed between depressive and no-depressive state against angry or sad speech \cite{koch2018neural}.

While depression has been consistently associated with reduced positive emotional reactivity \cite{Sheoran2022, Bylsma2008} favoring positive attenuation model of depression, the association with the negative emotional reactivity  have shown mixed results \cite{Sheoran2022, schlipf2013judgment, Beck2008}. Studies examining the emotional reaction against the negative stimuli have observed either no difference in emotional reactivity between depressive and no-depressive states for negative (e.g., angry or sad) stimuli \cite{koch2018neural, schlipf2013judgment} or reduced negative emotional reactivity to sad stimuli \cite{Bylsma2008, Rottenberg2017} or increased emotional reactivity to sad stimuli and blunted response for threatening stimuli by the individuals with depression \cite{kujawa2017vulnerability}, demanding complex comprehensive explanation. 

The Beck's cognitive model of depression \cite{Beck2008}, commonly known as the Cognitive Triad, argues that an individual internal representation determines how individuals perceive themselves and view the world around them and their future perspectives. During depressive episodes, an individual is assumed to favor processing negative \textit{compared to} positive stimuli, suggesting \textit{mood-congruent information processing} \cite{Beck2008, Bower1981}. In contrast, the emotional context insensitivity (ECI) model \cite{Rottenberg2017} argues that individuals with depressive episodes experience resistance to emotional alteration and fail to accommodate sudden environmental changes \cite{Bylsma2008, Rottenberg2017, Kuppens2010}. The two competing models offer alternative plausible explanation for the conflicting findings for emotional reactivity for negative stimuli \cite{Sheoran2022, Bylsma2008, koch2018neural, kujawa2017vulnerability}. In a nutshell, positive emotional reactivity offers a reliable measure for assessing emotional reactivity against positive stimuli, however, negative emotional reactivity demands further examination by controlling the methodological variability, like culture, participants' profile or stimuli type.

\begin{table*}[t]
    \caption{Descriptive Statistics for No-Depression \textbf{(ND)} and Depression \textbf{(D)} categories of participants across all emotions}
  \centering
  \small 
  \setlength{\tabcolsep}{4pt} 
  \begin{tabular}{@{}l|cccc|cccc|cccc|cccc|@{}}
    \toprule
    \multicolumn{1}{c|}{} &
    \multicolumn{16}{c|}{\textbf{Emotions}} \\
    \cmidrule(lr){2-5} \cmidrule(lr){6-9} \cmidrule(lr){10-13} \cmidrule(lr){14-17}
    & \multicolumn{4}{c|}{\textbf{Happy}} & \multicolumn{4}{c|}{\textbf{Sad}} & \multicolumn{4}{c|}{\textbf{Angry}} & \multicolumn{4}{c|}{\textbf{Neutral}} \\
    \cmidrule(lr){2-5} \cmidrule(lr){6-9} \cmidrule(lr){10-13} \cmidrule(lr){14-17}
    & \multicolumn{2}{c}{Val} & \multicolumn{2}{c|}{Arous} & \multicolumn{2}{c}{Val} & \multicolumn{2}{c|}{Arous} & \multicolumn{2}{c}{Val} & \multicolumn{2}{c|}{Arous} & \multicolumn{2}{c}{Val} & \multicolumn{2}{c|}{Arous} \\
    \cmidrule(lr){2-3} \cmidrule(lr){4-5} \cmidrule(lr){6-7} \cmidrule(lr){8-9} \cmidrule(lr){10-11} \cmidrule(lr){12-13} \cmidrule(lr){14-15} \cmidrule(lr){16-17}
    \hspace{2.8em}\textbf{N} & \makecell{Mean} & \makecell{SD} & \makecell{Mean} & \makecell{SD} & \makecell{Mean} & \makecell{SD} & \makecell{Mean} & \makecell{SD} & \makecell{Mean} & \makecell{SD} & \makecell{Mean} & \makecell{SD} & \makecell{Mean} & \makecell{SD} & \makecell{Mean} & \makecell{SD} \\
    \midrule
    \textbf{ND} \hspace{0.9em}(66) & 3.77 & 0.88 & 3.43 & 1.05 & 2.56 & 1.04 & 2.21 & 1.05 & 2.18 & 1.09 & 3.77 & 1.11 & 2.59 & 0.92 & 2.24 & 1.02 \\
    \textbf{D} \hspace{1.6em}(31) & 3.83 & 0.92 & 3.44 & 1.12 & 2.50 & 1.00 & 2.21 & 1.05 & 2.21 & 1.11 & 3.65 & 1.2 & 3.09 & 0.94 & 2.29 & 1.05\\

    \bottomrule
  \end{tabular}

  \label{tab:descriptive_stats}
\end{table*}

In our study, we examine the impact of depression on emotional reactivity to affective speech in which participants are asked to make judgement about the spekears' emotion while listening the speech audio recordings. We focus on college going young adults, chosen deliberately to examine the role of self-reported depression in emotion perception for a given speech audio recording. The choice of participants' profile for this study was determined by the reports \cite{liu2022influencing} stating the increasing prevalence of depression in college going young adults, especially after the post-pandemic episodes, and highlighted the debilitating impact of depression on students' academic, social and interpersonal lives. The current study will help us evaluate whether self-reported depressive symptoms exhibit similar trends in emotion perception of a particular speech.

\section{Methodology}

\subsection{Partcipants}
Total ninety seven (\textit{M} = 63, \textit{F} = 33, and \textit{O} = 1) university students, aged between 18-to-25 years volunteered for the study. All the participants reported normal hearing and high familiarity with English language. Their proficiency in listening American accented English was assessed by language fluency test, in which participants were made to listen to a paragraph with four questions. Participants were given chance to repeat the paragraph, until it is clear to them. 

At the time participation, minimal educational degree attained by the participants was at least higher secondary or 11th grade, with sixty one participants (62.8\%) reported pursing graduation degree, five (5.1\%) reported either pursuing or completed PhD degree and twenty three (23.7\%) indicated having completed Higher Secondary education, and 8 (8.2\%) participants reported having obtained a Post-Graduation degree. 

\begin{figure}[htbp]
    \centering
    \includegraphics[width=0.8\linewidth]{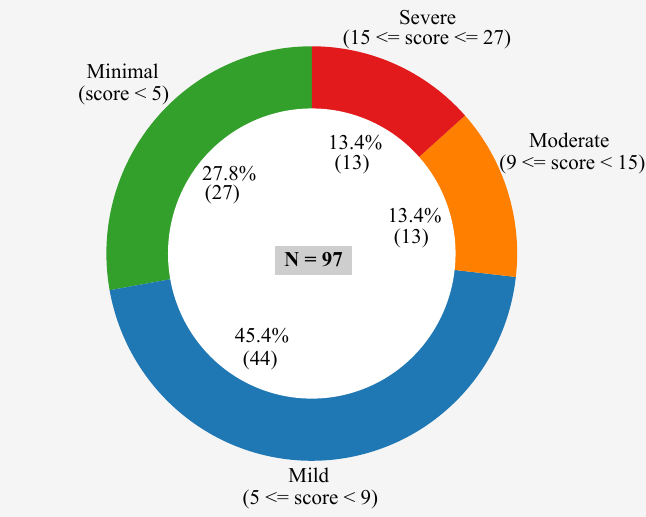}
    \caption{Percentage of Participants in different categories of Self-Reported Depression based on PHQ-9 Scores }
    \label{diagram1}
\end{figure}

\begin{figure*}[t!] 
  \centering
  \includegraphics[width=\textwidth]{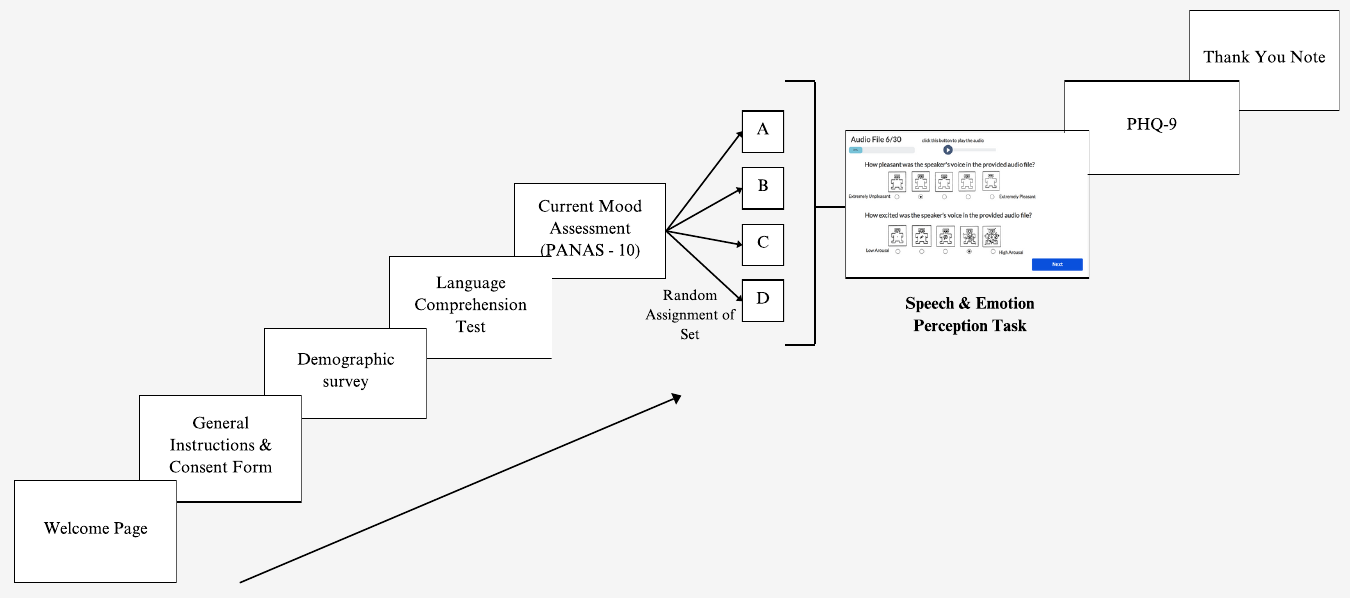}
  \caption{Overall schematic flow of the Experiment  }
  \label{fig:your_label}
\end{figure*}

\subsection{Stimuli } 

We used speech audio stimuli to elicit emotion. The speech stimuli were sourced from the IEMOCAP database \cite{Busso2008}, an open-source database created at the SAIL lab at USC, which contained dyadic sessions with actors performing scripted scenarios to elicit emotional expressions. This database is extensively annotated with categorical labels, such as anger, happiness, sadness, and neutrality, and dimensional annotations such as valence, activation, and dominance. 

We used both the categorical and dimensional approach to create four sets of stimuli, namely, happy, sad, angry, and neutral. The audio files were selected based on three criteria, i. the four categories, ii. the emotion rating using Likert scale 1-to-5, where 1= unpleasant/low arousal and 5=pleasant/high arousal, and iii. the similar duration of audio files, that ranged between 3-7 seconds. A total of 100 such audio files, 25 for each emotion category (sad, happy, angry, and neutral), were selected from the database \cite{Busso2008}. 

The descriptive Likert scale rating analysis of audio IEMOCAP databsed \cite{Busso2008} revealed distinct characteristics for each emotion category. Speech audio, labelled as sad emotion showed low valence (\textit{M}=2.16, \textit{SD}=2.16) and low arousal rating (\textit{M}=1.92, \textit{SD}= 0.81), whereas audio labelled as happy showed high valence (\textit{M}=4.04, \textit{SD}=0.35) and high arousal (\textit{M}= 3.12, \textit{SD} = 0.12). Audio labelled as angry emotions showed low valence and high arousal ratings, (\textit{M}= 1.48, \textit{SD} = 0.91; and \textit{M}=4.04, \textit{SD} = 0.84, respectively), while neutral emotions comprised medium valence and arousal, (\textit{M} = 3.12, \textit{SD}= 0.33; and \textit{M}=2.36, \textit{SD}= 0.57, respectively).  

To mitigate any potential biases introduced by the specific content of audio stimuli, we employed a pseudo-randomized design by creating four distinct stimulus sets labeled A, B, C, and D. Each set consisted of 30 audio files (25 unique and 5 common audio files across sets). We included almost an equal number of files from each emotion in each set ensuring a balanced representation of all emotions. For instance, in set A, there are 6 files each from Sad, Angry and Happy emotions, and 7 files from Neutral) and in each set one of the four emotions has 7 files whereas the rest has 6. This approach improved the internal validity of the study by minimizing the influence of any idiosyncrasies associated with a particular set of stimuli, thus allowing for more robust generalizations across a diverse range of emotional expressions. We ensured that in each set no 2 files labeled with the same emotion are consecutively present. 

Participants were randomly assigned to one of four sets (A, B, C, D). Set A was allotted to 23 participants, Set B to 26, Set C to 24, and Set D to 24 participants. To ensure intra-rater reliability, we included duplicates of the first 5 files at the end of each set. This resulted in each set containing a total of 30 files for participants to perceive and rate. 

\subsection{Tasks}
Each participant performed a total set of three main tasks, comprising a patient healthy questionnaire (PHQ-9), positive and negative scale (PANAS-10), and affective speech perception task, alongside demographic survey, which included information related to their personal and familial health history, gender identity, and hearing ability. We also performed the language fluency test for the English language to reduce the confound caused by language. The description of the affective speech perception task, PHQ-9, PANAS and the language fluency test as follow:

\begin{description}
    
\vspace{0.3em}
\item[(1) Emotion Perception Task:] In this task each participants were presented with a speech audio file, depicting either  happy, sad, angry, or neutral emotion. Participants were instructed to listen to the audio file, and judge the speakers' emotion on valence and arousal scale, using 5 point Likert SAM scale, where 1 = extremely unpleasant or low arousal, and 5 = extremely pleasant/ high arousal. The audio file was presented until both responses were made, and participants were allowed to repeat the audio if required (Figure 2, refer to the interface box). There were 30 such files were presented. The ratings for both valence and arousal were logged separately for each file. We also recorded total time taken, as a response time, to analyze the latency in emotional perceptual processing across states of depression. Response time was measured from the onset of audio stimuli to participants' click to move to the next stimuli. The timing reset for each subsequent stimulus. 


\item [(2) Positive and Negative Scale (PANAS)] We used Positive and Negative Affect Scale (PANAS-10) , originally developed by Watson and Clark \cite{watson1992panas} , to assess current mood state. PANAS 10 comprises of 10 specific affects: upset, hostile, alert, ashamed, inspired, nervous, determined, attentive, afraid, and active. Of which, five affects are positive and rest five are of negative emotion. Participants were instructed to carefully read the items and report their current feeling on a five point scale, where 1 referred to very slightly or not at all and 5 referred to extremely. The score is calculated by summing up the total positive and negative items separately. For positive items, high score represents high levels of positive effect, and for the negative items, low score represents low levels of negative effect. In other words, the high positive score and low negative score in PANAS signifies healthy scores. 

\item[(3) Patient Health Questionnaire (PHQ-9)] 
 
The PHQ-9 is developed by Kroenke and colleagues, \cite{kroenke2001phq}  to assess the individual experience of depression, over the preceding two weeks, including the date of assessment. It comprises statement related to cognitive, emotional, and somatic experiences, such as Over the last 2 weeks, \textit{how often have you been bothered by the following problems?} For example, \textit{Little interest or pleasure in doing things}. Participants were instructed to read the statements carefully and choose the options that describe their feeling most appropriately on a four point Likert scale, where 0 referred to not at all, and 3 referred to nearly every day. The total score, ranging from 0 to 27, is obtained by summing the ratings across all nine items. The cut-off scores for categorizing depression severity were set at 5, 10, 15, and 20, representing mild, moderate, moderately severe, and severe depression, respectively (Figure 1). For statistical analysis, we used two groups, \textit{no-depression} comprising minimal and mild category, and \textit{depression} comprising moderate and above categories (Figure 2 and 3).
\item [(4) Language Fluency Test] A comprehension test was administered to ensure participants' ability to understand American accent audio files used in the perception task. It included listening to two audio files followed by four simple multiple-choice questions assessing comprehension. The cumulative score determined participants' proficiency, with an average total fluency score of 5.50/7 (SD = 0.7) across all participants, establishing eligibility for the subsequent speech perception task.
    \vspace{0.3em}

\end{description}

\subsection{Procedure} 
The study involved participants completing a series of tasks involving surveys and emotion perception task after listening to speech audio recordings. The study was conducted using the software Labvanced \cite{finger2017labvanced}, a JavaScript web application designed for professional behavioral research. 
Data collection was performed in a controlled environment to ensure consistency across participant experiences. Participants were seated in a noise-proof room equipped with a comfortable chair and a computer workstation. To minimize external distractions, each participant used a high-quality headset to listen to the audio stimuli. The room was soundproofed to minimize interference from external noise.
An instructor was present in an adjacent room to assist participants as needed and to monitor the overall progress of the study. The instructor was available to provide clarifications and address any concerns that participants might have encountered during the tasks.

Each session began with a welcome note and a brief introduction of the study. Upon agreeing to participate, participants were asked to read the consent form, which informed them about their rights to privacy and withdrawal of their participation without any future consequences. To ensure compliance with ethical standards, participants digitally acknowledged their consent before commencing the study. 
Subsequently, they received detailed instructions from the instructor and on-screen prompts outlining the study's procedures. Participants were briefed on the tasks without explicitly disclosing the research objectives, thus mitigating potential demand characteristics (Figure 2).



\begin{figure*}[t!] 
  \centering
  \includegraphics[width=\textwidth]{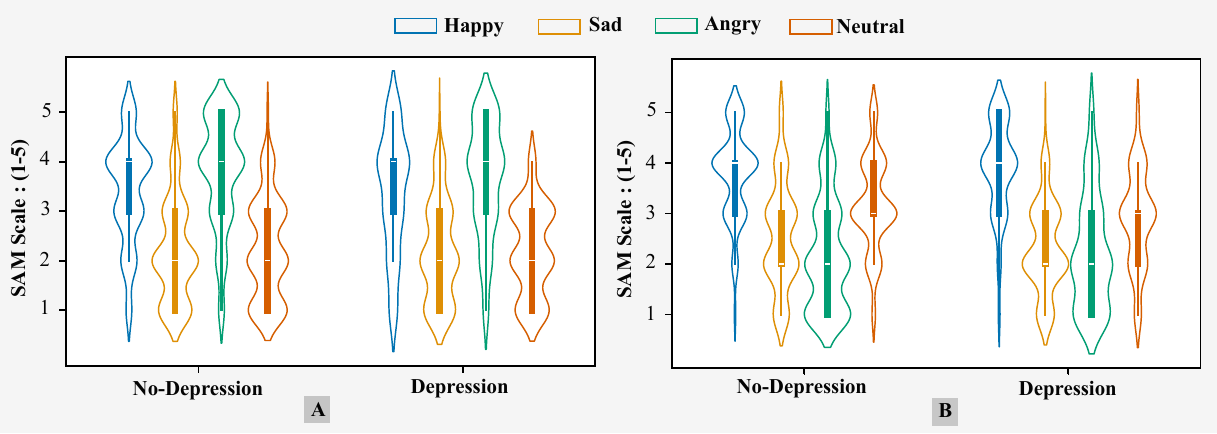}
  \caption{\textbf{(A)} - Arousal and \textbf{(B)} - Valence Ratings by No-Depression and Depression categories of participants across 4 emotions}
  \label{fig:your_label}
\end{figure*}



\section{Statistical Analyses}

The current study examined the role of self-reported depression in emotional reaction to the affective speech audio information. We performed the power analysis using G*Power version 3.1 \cite{faul2007g} to determine the minimum number of participants required to test the study hypothesis. We obtained sample size, N= 36, with 95\% power, a moderate effect size (.25) in an ANOVA repeated measure, within-between interaction. Given that the calculated power analysis did not account of post-hoc tests that are be required for the multiple comparisons, we revised the sample size calculation by multiplying it with 4.1 (Brooks and Johanson, 2011), and obtained n=144 with a revised minimal sample size required to test the hypothesis. 

We are presenting a preliminary data of n=97 participants, with unequal sample size data, of which n=71 reported minimal and mild self-reported depressive states ($<$ 9 scores on PHQ-9, Figure 1), equivalent to healthy adults, i.e., no-depression group (ND), and n=26 reported moderate or severe experience of depressive states ($>$/$=$ 9 on PHQ-9 survey, Figure 1) considered unhealthy, i.e., depression (D) group. The Shapiro-Wilk test showed that the current data violated the normality assumption, (\textit{p} $<$ .05, and lead us to use non-parametric comparative analysis, using Mann-Whitney test, with a Bonferroni correction (\textit{p} $<$ 0.012) to avoid error caused by multiple comparisons. We performed three separate analyses for valence, arousal and reaction time across no-depression (ND) and  (D) groups.

\section{Results and Discussion}

For valence rating, we observed a significant effect of self-reported depression on perception of affective speech depicting neutral emotion with a medium effect size (\textit{U} = 570.5, \textit{p} = .002, \textit{r} = -0.382). We observed a significnatly lower rating (\textit{Med} = 2.5) by individuals with depressive than no-depressive state (3.0).  However, the comparison between ND and D with other stimuli depicting happy, sad, or angry did not yield any significant difference (Figure 3). 

Further, ND and D did not show any significant difference in arousal ratings (Table 1) and response time while processing audio files depicting happy, sad, angry, and neutral emotion. 

We further analyzed the effect of predisposed mood on current mood, positive and negative scale (PANAS), and observed a significantly higher PANAS-negative score (\textit{Mdn} = 8.5) by individuals with self-reported depression than no-depression (\textit{Mdn} = 6.0), (\textit{U} = 1382.5, \textit{p} $<$ 0.001, \textit{r} = .498) (Figure 4). 

Contrary to previous findings, we did not observe a significant difference in positive speech perception between individuals with and without depression \cite{uekermann2008perception, Sheoran2022, Bylsma2008,   koch2018neural, schlipf2013judgment}. However, our results demonstrated consistent emotional reactivity for speech stimuli depicting sadness and anger \cite{uekermann2008perception, koch2018neural}. This result was consistent across three critical measures of emotion perception, namely valence, arousal, and reaction time, except the valence rating for neutral stimuli. 

The absence of a difference in positive affective rating challenges the previous finding with reduced positive emotional reactivity to positive stimuli. However, the failure to reject the null hypothesis across all emotional categories raises the possibility that the Americanized automated speech audio files may have influenced the results, emphasizing the need for culturally appropriate stimuli. While participants reported high familiarity with the audio files, the Americanized accented speech might have demanded a focus shift to the content and overshadowed the emotional aspect of speech perception. Future studies should consider employing Indian language or Indianized English speech audio files to evaluate emotional reactivity within distinct cultural contexts comprehensively.


\begin{figure}[htbp]
    \centering
    \includegraphics[width=1.0\linewidth]{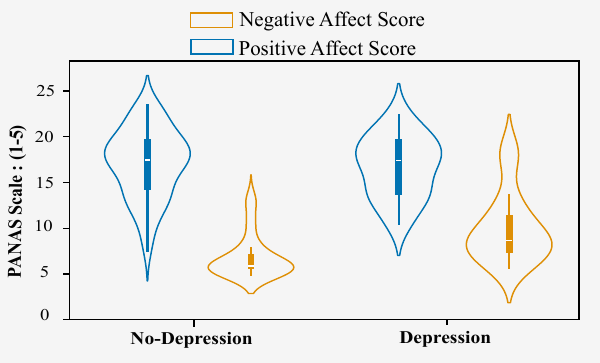}
    \caption{PANAS Scores for No-Depression and Depression category of Participants }
    \label{diagram1}
\end{figure}

\section{Conclusion}
The current study examines the relationship between depression and the perception of affective speech stimuli by asking participants to judge the speaker's emotion on valence and arousal. We observed no significant effects of depression on emotional perception except the alteration of valence responses to neutral stimuli. Notably, our findings contradict previous research by revealing no reduced positive emotional reactivity in individuals with depression. Despite acknowledging limitations, such as reliance on self-reported depression without clinical assessment and the use of Americanized English speech audio files, our study contributes to the emotional reactivity literature within the sub-clinical population. Most importantly, no effect of depression on negative emotional reactivity aligns with both behavioural and neural correlates data of patients diagnosed with affective disorders or depression. The findings suggest future research addressing methodological considerations and cultural nuances in emotional reactivity assessment.

\bibliographystyle{IEEEtran}
\bibliography{mybib}

\end{document}